\def\year{2020}\relax
\documentclass[letterpaper]{article} 
\usepackage{aaai20}  
\usepackage{times}  
\usepackage{helvet} 
\usepackage{courier}  
\usepackage[hyphens]{url}  
\usepackage{graphicx} 
\usepackage{graphicx}  
\usepackage{calligra}
\usepackage{calrsfs}
\usepackage{paralist, tabularx}
\usepackage{algorithm}
\usepackage{algorithmic}
\usepackage{amssymb}
\usepackage{amsmath}
\usepackage{commath}
\usepackage{booktabs}
\usepackage{multirow}
\usepackage{color}
\def \z {z}

\def \c {\mathbf{c}}

\def \x {\mathbf{x}}
\def \y {\mathbf{y}}
\def \c {\mathbf{c}}
\def \X {\mathbf{X}}
\def \Y {\mathbf{Y}}
\def \h {\mathbf{h}}
\def \s {s}
\DeclareMathOperator*{\argmin}{arg\,min}
\DeclareMathOperator*{\argmax}{arg\,max}
\newcommand{\bdelta}{\boldsymbol{\delta}}
\newcommand{\bx}{\mathbf{x}}
\newcommand{\bw}{\mathbf{w}}

\newcommand{\citet}[1]{\citeauthor{#1}\shortcite{#1}}
\newcommand{\citep}{\cite}

\title{Seq2Sick: Evaluating the Robustness of Sequence-to-Sequence Models with Adversarial Examples}
\author{Minhao Cheng,\textsuperscript{\rm 1} Jinfeng Yi,\textsuperscript{\rm 2} Pin-Yu Chen,\textsuperscript{\rm 3} Huan Zhang,\textsuperscript{\rm 1} Cho-Jui Hsieh\textsuperscript{\rm 1}\\ 
\textsuperscript{\rm 1}Department of Computer Science, UCLA, \textsuperscript{\rm 2}JD AI Research,\textsuperscript{\rm 3}IBM Research\\ 
\{mhcheng,huanzhang,chohsieh\}@cs.ucla.edu, yijinfeng@jd.com, pin-yu.chen@ibm.com  
}
 
\begin{document}

\maketitle

\begin{abstract}
Crafting adversarial examples has become an important technique to evaluate the robustness of deep neural networks (DNNs). 
However, most existing works focus on attacking the image classification problem since its input space is continuous and output space is finite. 
 In this paper, we study the much more challenging problem of crafting adversarial examples for sequence-to-sequence (seq2seq) models, whose inputs are discrete text strings and outputs have an almost infinite number of possibilities. To address the challenges caused by the discrete input space, we propose a projected gradient method combined with group lasso and gradient regularization. To handle the almost infinite output space, we design some novel loss functions to conduct non-overlapping attack and targeted keyword attack. We apply our algorithm to machine translation and text summarization tasks, and verify the effectiveness of the proposed algorithm: by changing less than 3 words, we can make seq2seq model to produce desired outputs with high success rates. We also use an external sentiment classifier to verify the property of preserving semantic meanings for our generated adversarial examples.
 On the other hand, we recognize that, compared with the well-evaluated CNN-based classifiers, seq2seq models are intrinsically more robust to adversarial attacks. 
\end{abstract}

\section{Introduction}

Adversarial attack on deep neural networks (DNNs) aims to slightly modify the inputs of DNNs and mislead them to make wrong predictions~\cite{DBLP:journals/corr/SzegedyZSBEGF13,goodfellow2014explaining}. This task has become a common approach to evaluate the robustness of DNNs 
-- generally speaking, the easier an adversarial example can be generated, the less robust the DNN model is. However, models designed for different tasks are not born equal: some tasks are strictly harder to attack than others. For example, attacking an image is much easier than attacking a text string, since image space is continuous and the adversary can make arbitrarily small changes to the input. Therefore, even if most of the pixels of an image have been modified, the perturbations can still be imperceptible to humans when the accumulated distortion is small. In contrast, text strings live in a discrete space, and word-level manipulations may significantly change the meaning of the text. In this scenario,  an adversary should change as few words as possible, and hence this limitation induces a sparse constraint on word-level changes. Likewise, attacking a classifier should also be much easier than attacking a model with sequence outputs. This is because different from the classification problem that has a finite set of discrete class labels, the output space of sequences may have an almost infinite number of
possibilities. If we treat each sequence as a label, a targeted attack needs to find a specific one over an enormous number of possible labels, leading to a nearly zero volume in search space. 
This may explain why most existing works on adversarial attack focus on the image classification task, since its input space is continuous and its output space is finite.

In this paper, we study a harder problem of crafting adversarial examples for sequence-to-sequence (seq2seq) models~\cite{DBLP:conf/nips/SutskeverVL14}. This problem is challenging since it combines both aforementioned difficulties, i.e., discrete inputs and sequence outputs with an almost infinite number of possibilities. We choose this problem not only because it is challenging, but also because seq2seq models are widely used in many safety and security sensitive applications, e.g., machine translation~\citep{bahdanau2014neural}, 
text summarization~\citep{rush2015neural}, 
and speech recognition~\cite{chan2016listen}, thus measuring its robustness becomes critical. Specifically, we aim to examine
the following questions in this study:
\begin{enumerate}
\item \emph{Is it possible to slightly modify the inputs of seq2seq models while significantly change their outputs?}
\item \emph{Are seq2seq models more robust than the well-evaluated CNN-based image classifiers?}
\end{enumerate}

We provide an affirmative answer to the first question by developing an effective adversarial attack framework called Seq2Sick. 
It is an optimization-based framework that aims to learn an input sequence that is close enough to the original sequence {(in terms of distance metrics in word embedding spaces or sentiment classification)} while leads to the desired outputs with high confidence. To address the challenges caused by the discrete input space, we propose to use the projected gradient descent method combined with group lasso and gradient regularization. To address the challenges of almost infinite output space, we design some novel loss functions for the tasks of non-overlapping attack and targeted keyword attack. Our experimental results show that the proposed framework yields high success rates in both tasks. 
However, even if the proposed approach can successfully attack seq2seq models, our answer to the second question is ``Yes''. Compared with CNN-based classifiers that are highly sensitive to adversarial examples, seq2seq model is intrinsically more robust since it has discrete input space and the output space is exponentially large. As a result, adversarial examples of seq2seq models usually have larger distortions and are more perceptible than the adversarial examples crafted for CNN-based image classifiers. {To the best of our knowledge, this paper is the first work that evaluates the robustness of seq2seq model, which has inspired many follow-up works and has been cited since its debut.}

\section{Related work and Background}
\citet{papernot2016crafting} first uses Fast Gradient Sign Method (FGSM) to conduct an attack on RNN/LSTM-based classification problems. 
In order to generate text adversarial examples,~\citet{li2016understanding} proposes to use reinforcement learning to locate important words that could be deleted in sentiment classification.  \citet{samanta2017towards} and \citet{liang2017deep} generate adversarial sequences by inserting or replacing existing words with typos and synonyms. 
\citet{gao2018black} aims to attack sentiment classification models in a black-box setting. It develops some scoring functions to find the most
important words to modify. \citet{yang2018greedy} applied a greedy approach and a Gumbel trick to speed up the inference time. \citet{alzantot2018generating} proposed a genetic algorithm to attack sentiment analysis. 
These approaches differ from our method in that they study simple text classification problems while we focus on the more challenging seq2seq model with sequential outputs. Other than attacking text classifiers, \citet{jia2017adversarial}  aims to fool reading comprehension systems by adding misleading sentences, which has a different focus than ours. \citet{zhao2017generating} uses the generative adversarial network (GAN) to craft natural adversarial examples.  However, it can only perform the untargeted attack and also suffers from high computational cost. 

\begin{table*}[htbp]
\caption{Summary of existing works that are designed to attack RNN models. ``BINARY'' indicates the attack is for binary classifications, and there is no difference between untargeted and targeted attack in this case. ``CLASS'' means targeted attack to a specific class. ``KEYWORD'' means targeted attack to a specific keyword. Here we omit follow-up works based on Seq2Sick.}
\label{tb:ref}
\begin{center}
\begin{small}
\begin{tabular}{lccccr}
\toprule
Methods &\! Gradient Based? \!&\! Word-level RNN?\! &\! Sequential Output?\! &\! Targeted Attack?\! \\
\midrule
\citet{ebrahimi2017hotflip} & $\surd$ & $\times$ & $\surd$ & Class\\
\citet{jia2017adversarial} & $\times$ & $\surd$ & $\times$ & $\times$ \\
\citet{li2016understanding} & $\surd$ & $\surd$ & $\times$ & Class \\
\citet{papernot2016crafting} & $\surd$ & $\times$ & $\surd$ & $\times$ \\
\citet{gao2018black} & $\times$ & $\surd$ & $\times$ & Binary \\
\citet{samanta2017towards} & $\times$ & $\times$ & $\times$ & Binary \\
\citet{zhao2017generating} & / & $\surd$ & $\surd$ & Class \\
\citet{liang2017deep} & $\surd$ & $\times$ & $\times$ &Class \\ 
\citet{alzantot2018generating} & $\times$ & $\surd$ & $\times$ &Class \\ 
\citet{yang2018greedy} & $\times$ & $\surd$ & $\times$ &Class \\
Seq2Sick (Ours) & $\surd$ &$\surd$&$\surd$& Keyword\\ 
\bottomrule
\end{tabular}
\end{small}
\end{center}
\end{table*}

Notably, almost all the previous methods are based on greedy search, i.e., at each step, they search for the best word and the best position to replace the previous word. As a result, 
their search space grows rapidly as the length of input sequence increases. To address this issue, we propose a novel approach that uses group lasso regularization and the projected gradient descent method with gradient regularization to simultaneously search all the replacement positions. Table~\ref{tb:ref} summarizes the key differences between the proposed framework Seq2Sick and the existing attack methods on RNN-based models.
Note that our paper was the first method for attacking seq2seq model on arXiv and after our work, there are some followup papers such as~\cite{michel2019evaluation}, where they use several similarity metrics to conduct the attack while our work are focusing on the BLEU score and self-defined loss functions.

Before introducing the proposed algorithms, we first briefly describe the sequence-to-sequence (seq2seq) model. 
Let $\x_i\in \mathbb{R}^d$ be the embedding vector of each input word,
$N$ be the input sequence length, and $M$ be the output sequence length. Let $\omega$ be the input vocabulary, and the output word $\y_j\in \nu $ where $\nu$ is the output vocabulary.  
The seq2seq model has an encoder-decoder framework that aims at mapping an input sequence of vectors 
$\X=(\x_1,\dots,\x_N)$ to the output sequence $\Y=\{\y_1,\dots,\y_{M}\}$. 
Its encoder first reads the input sequence, then each RNN/LSTM cell computes $\h_t=f(\x_t,\ \h_{t-1})$, where $\x_t$ is the current input, $\h_{t-1}$ and $\h_{t}$ represent the previous and current cells' hidden states, respectively. The next step computes the context vector $\c$ using all the hidden layers of cells $\h_1,\dots,\h_N$, i.e $\c = q({\h_1,\cdots,\h_N})$, where $q(\cdot)$ could be a linear or non-linear function. In this paper, we follow the setting in ~\citep{DBLP:conf/nips/SutskeverVL14} that $\c = q({\h_1,\cdots,\h_N}) = \h_N$. 

Given the context vector $\c$ and all the previously words $\{\y_1,\dots,\y_{t-1}\}$, the decoder is trained to predict the next word $\y_{t}$. 
Specifically, the $t$-th cell in the decoder receives its previous cell's output $\y_{t-1}$ and the context vector $\c$, and then outputs
 \begin{equation}
     \z_t = g(\y_{t-1},\c)\ \  \text{and}\ \ p_t = \text{softmax}(\z_t), 
 \end{equation}
 where $g$ is another RNN/LSTM cell function.  $\z_t:=[z_t^{(1)},z_t^{(2)},\dots,z_t^{(|\nu|)}] \in \mathbb{R}^{|\nu|}$ is a vector of the \emph{logits} for each possible word in the output vocabulary $\nu$. 

\section{Seq2Sick: Proposed Framework}
Crafting adversarial examples against the seq2seq model can be formulated as an optimization problem:
\begin{align}
 \min\nolimits_{\bdelta}\ L(\X+\bdelta) + \lambda \cdot R(\bdelta),
 \label{eq:framework}
\end{align}
where 
$R(\cdot)$ indicates the regularization function to measure the magnitude of distortions. $L(\cdot)$ is the loss function to penalize the unsuccessful attack and it may take different forms in different attack scenarios. A common choice for $R(\bdelta)$ is the $\ell_2$ penalty $\norm{\bdelta}_2^2$, but it is, as we will show later, not suitable for attacking seq2seq model. $\lambda>0$ is the regularization parameter that balances the distortion and attack success rate -- a smaller $\lambda$ will make the attack more likely to succeed but with the price of larger distortion. 

In this work, we focus on two kinds of attacks: \textit{non-overlapping attack} and \textit{targeted keywords attack}. The first attack requires that the output of the adversarial example shares no overlapping words with the original output. This task is strictly harder than untargeted attack, which only requires that the adversarial output to be different from the original output \cite{zhao2017generating,ebrahimi2017hotflip}. We ignore the task of untargeted attack since it is trivial for the proposed framework, which can easily achieve a 100\% attack success rate, while \citet{ebrahimi2017hotflip} could achieve 76.24\% attack success rate for text summarization and  98.8\% success rate for machine translation with 1 word change. Targeted keywords attack is an even more challenging task than non-overlapping attack. Given a set of targeted keywords, the goal of targeted keywords attack is to find an adversarial input sequence such that all the keywords must appear in its corresponding output. In the following, we respectively introduce the loss functions developed for the two attack approaches.

\subsubsection{\textbf{Non-overlapping Attack}}
To formally define the non-overlapping attack, we let 
$\mathbf{s}=\{\s_1,\dots,\s_M\}$ be the original output sequence,  where $s_i$ denotes the location of the $i$-th word in the output vocabulary $\nu$. $\{\z_1, \dots, \z_M\}$ 
indicates the logit layer outputs of the adversarial example. 
In the non-overlapping attack, the output of adversarial example should be entirely different from the original output $\mathbf{S}$, i.e., 
\begin{equation*}
    \s_t \neq \argmax\nolimits_{y \in \nu}\z_t^{(y)}, \ \ \forall t=1, \dots, M,
\end{equation*}
which is equivalent to
 \begin{equation*}
  \z_t^{(s_t)} < \max\nolimits_{y \in \nu,\ y\neq s_t}\z_t^{(y)}, \ \ \forall t=1, \dots, M.    
 \end{equation*}
Given this observation, we can define a hinge-like loss function $L$ to generate adversarial examples in the non-overlapping attack, i.e., 
\begin{equation}
L_{\text{non-overlapping}} = \sum\nolimits_{t=1}^M \max\{-\epsilon,\ \z_t^{(s_t)}-\max_{y\neq s_t}\{\z_t^{(y)}\}\},
\label{eq:untargeted}
\end{equation}
where $\epsilon\ge 0$ denotes the confidence margin parameter. Generally speaking, a larger $\epsilon$ will lead to a more confident output and a higher success rate, but with the cost of more iterations and longer running time.

We note that non-overlapping attack is much more challenging than untargeted attack, which suffices to find a one-word difference from the original output~\citep{zhao2017generating,ebrahimi2017hotflip}. We do not take untargeted attack into account since it is straightforward and the replaced words could be some less important words such as ``the'' and ``a''. 

\subsubsection{\textbf{Targeted Keywords Attack}}
Given a set of targeted keywords, the goal of targeted keywords attack is to generate an adversarial input sequence to ensure that all the targeted keywords appear in the output sequence. This task is important since it suggests adding a few malicious keywords can completely change the meaning of the output sequence. For example, in  English to German translation, an input sentence ``\emph{policeman helps protesters to keep the assembly in order}'' should generate an output sentence ``\emph{Polizist hilft Demonstranten, die Versammlung in Ordnung zu halten}''. However, changing only one word from ``hilft'' to ``verhaftet'' in the output will significantly change its meaning, as the new sentence means ``\emph{police officer arrested protesters to keep the assembly in order}''. 

In our method, we do not specify the positions of the targeted keywords in the output sentence. Instead, it is more natural to design a loss function that allows the targeted keywords to become
the top-1 prediction at any positions. The attack is considered as successful only when ALL the targeted keywords appear in the output sequence. Therefore, the more targeted keywords there are, the harder the attack is. To illustrate our method, we start from the simpler case with only one targeted keyword $k_1$. To ensure that the target keyword word's logit $z^{(k_1)}_t$ be the largest among all the words at a position $t$, we design the following loss function:
\begin{equation}
L = \min_{t \in[M]}\{\max\{-\epsilon,\ \max_{y\neq k_1}\{z_t^{(y)}\}-z_t^{(k_1)}\}\},
\end{equation}
which essentially searches the minimum of the hinge-like loss terms over all the possible locations $t\in [M]$. 
When there exist more than one targeted keywords $K = \{k_1,k_2,\dots,k_{|K|}\}$, where $k_i$ denotes the $i$-th word in output vocabulary $\nu$, we follow the same idea to define the loss function as follows:
\begin{equation}
L_{\text{keywords}} = \sum_{i=1}^{|K|} \min_{t \in[M]}\{\max\{-\epsilon,\max_{y\neq k_i}\{z_t^{(y)}\}-z_t^{(k_i)}\}\}.
\label{eq:loss_keyword_1}
\end{equation}
However, the loss defined in~\eqref{eq:loss_keyword_1} suffers from the ``keyword collision'' problem. When there are more than one keyword,  it is possible that multiple keywords compete at the same position to attack. 
To address this issue, we define a mask function $m$ to mask off the position if it has been already occupied by one of the targeted keywords:
\begin{equation}
    m_{t}(x) = 
    \begin{cases}
    +\infty & \text{if } \argmax_{i\in \nu} z_t^{(i)} \in K\\
    x            & \text{otherwise}
    \end{cases}
\end{equation}
In other words, if any of the keywords appear at position $t$ as the top-1 word, we ignore that position and only consider other positions for the placement of remaining keywords.
By incorporating the mask function, the final loss for targeted keyword attack becomes:
\begin{equation}
\sum_{i=1}^{|K|} \min_{t \in[M]}\{m_{t}(\max\{-\epsilon,\ \max_{y\neq k_i}\{z_t^{(y)}\}-z_t^{(k_i)}\})\}.   \label{eq:keyword} 
\end{equation}

\subsection{Handling Discrete Input Space}
As mentioned before, the problem of ``discrete input space'' is one of the major challenges in attacking seq2seq model. Let $\mathbb{W}$ be the set of word embeddings of all words in the input vocabulary. A naive approach is to first learn $\X+\bdelta^*$ in the continuous space by solving the problem~\eqref{eq:framework}, 
and then search for its nearest word embedding in $\mathbb{W}$. 
This idea has been used in attacking sequence classification models in~\citet{gong2018adversarial}. Unfortunately, when applying this idea to targeted keywords attack, we report that all of the 100 attacked sequences on Gigaword dataset failed to generate the targeted keywords.
The main reason is that by directly solving~\eqref{eq:framework}, the final solution will not be a  feasible word embedding in $\mathbb{W}$, and its nearest neighbor could be far away from it due to the curse of dimensionality~\cite{friedman1997bias}.
 
To address this issue, we propose to add an additional constraint to enforce that $\X+\bdelta$ belongs to the input vocabulary $\mathbb{W}$. The optimization problem then becomes
\begin{equation}
\begin{aligned}
\min_{\bdelta}&\quad L(\X+\bdelta) + \lambda \cdot R(\bdelta)\\
\text{s.t.}& \quad  \x_i+\delta_i \in \mathbb{W}\quad \forall i=1,\dots,N\\
\end{aligned}
\end{equation}
We then apply  projected gradient descent to solve this constrained problem. At each iteration, we project the current solution $\x_i+\delta_i$, where $\delta_i$ denotes the $i$-th column of $\bdelta$, back into $\mathbb{W}$ to ensure that $\X+\bdelta$ can map to a specific input word.

\paragraph{Group lasso Regularization:}
$\ell_2$ norm has been widely used in the adversarial machine
learning literature to measure distortions. 
However, it is not suitable for our task since almost all the learned $\{\delta_t\}_{t=1}^M$ using $\ell_2$ regularization will be nonzero. As a result, most of the inputs words will be perturbed to another word, leading to an adversarial sequence that is significantly different from the input sequence.

To solve this problem, we treat each $\delta_t$ with $d$ variables as a group, and use the group lasso regularization 

\begin{equation*}
    R(\bdelta) = \sum\nolimits_{t=1}^N \|\delta_t\|_2 
\end{equation*}

to enforce the group sparsity: only a few groups (words) in the optimal solution $\bdelta^*$ are allowed to be nonzero. 
\subsection{Gradient Regularization}

When attacking the seq2seq model, it is common to find that the adversarial example is located in a region with very few or even no embedding vector. This will negatively affect our projected gradient method since even the closest embedding from those regions can be far away. 

To address this issue, we propose a gradient regularization to make $\X+\bdelta$ close to the word embedding space. 
Our final objective function becomes:
\begin{align}
\min_{\bdelta}\quad \!\!\!\!\!&L(\X\!+\!\bdelta) \!+\! \lambda_1 \sum_{i=1}^N \norm{\delta_i}_2 
\!+\!\lambda_2 \sum_{i=1}^N \min_{\bw_j\in \mathbb{W}}\{\norm{\x_i+\delta_i-\bw_j}_2\}\nonumber\\
\text{s.t. } &  \x_i+\delta_i \in \mathbb{W}\quad \forall i=1,\dots,N
\label{eq:loss}
\end{align}
where the third term is our gradient regularization that penalizes a large distance to the nearest point in $\mathbb{W}$. The gradient of this term can be efficiently computed since it is only related to one $\bw_j$ that has a minimum distance from $\x_i+\delta_i$. 
For the other terms, we use the proximal operator to optimize the group lasso regularization, and the gradient of the loss function $L$ can be computed through back-propagation. The detailed steps of our approach, Seq2Sick, is presented in Algorithm~\ref{alg:opt}. Our source code is publicly available at \url{https://github.com/cmhcbb/Seq2Sick}. 

\paragraph{Computational Cost:} Our algorithm needs only one back-propagation to compute the gradient $\nabla_{\bdelta}L(x+\bdelta)$. The bottleneck here is to project the solution back into the word embedding space, which depends on the number of words in the input dictionary of the model.~\citet{gong2018adversarial} uses $GloVe$ word embedding~\cite{pennington2014glove} that contains millions of words to do a nearest neighbor search. Fortunately, our model does not need to use any pre-trained word embedding, thus making it a more generic attack that does not depend on pre-trained word embedding. 
Besides, we can employ approximate nearest neighbor (ANN)  approaches to further speed up the projection step.
\begin{algorithm}[tb]
   \caption{Seq2Sick algorithm}
   \label{alg:opt}
\begin{algorithmic}
   \STATE {\bfseries Input:} input sequence $\bx=\{x_1,\dots,x_N\}$, seq2seq model, target keyword $\{k_1,\dots,k_T\}$
   \STATE {\bfseries Output:} adversarial sequence $\bx^*=\bx + \bdelta^*$
   \STATE Let $\mathbf{s}=\{s_1,\dots,s_M\}$ denote the original output of $\bx$.
   \STATE Set the loss $L(\cdot)$ in~\eqref{eq:loss} to be ~\eqref{eq:untargeted} 
   \IF{\ Targeted Keyword Attack}
   \STATE Set the loss $L(\cdot)$ in~\eqref{eq:loss} to be ~\eqref{eq:keyword}
   \ENDIF
   \FOR{$r=1,2,\dots, T$}
   \STATE back-propagation $L$ to achieve gradient $\nabla_{\bdelta}L(\bx+\bdelta_r)$
   \FOR{$i=1,2,\dots,N$}
   \IF{$\norm{\delta_{r,i}} > \eta\lambda_1$}
   \STATE $\delta_{r,i} = \delta_{r,i} - \eta\lambda_1 \frac{\delta_{r,i}}{\norm{\delta_{r,i}}}$
   \ELSE   
   \STATE $\delta_{r,i} = 0$
   \ENDIF
   \ENDFOR
   \STATE $y^{r+1} = \bdelta^{r} + \eta \cdot \nabla_{\bdelta}L(\bx+\bdelta^r)$
   \STATE $\bdelta^{r+1} =\argmin\limits_{\bx+\bdelta^{r+1} \in \mathbb{W}}\norm{y^{r+1}-\bdelta^{r+1}}$
   \ENDFOR
   \STATE $\bdelta^* = \bdelta^T$
   \STATE $\bx^* = \bx + \bdelta^*$
   \STATE {\bfseries return} $\bx^*$
\end{algorithmic}
\end{algorithm}

\section{Experiments}
We conduct experiments on two widely-used applications of seq2seq model: text summarization and machine translation. 
\subsection{Datasets}
We use three datasets DUC2003, DUC2004
, and Gigaword
, to conduct our attack for the text summarization task. Among them, DUC2003 and DUC2004 are widely-used datasets in documentation summarization. We also include a subset of randomly chosen samples from Gigaword to further evaluate the performance of our algorithm. For the machine translation task, we use 500 samples from WMT'16 Multimodal Translation task. The statistics about the datasets are shown in Table ~\ref{tb:datsets}. 
\begin{table}[htbp]
\caption{Statistics of the datasets. ``\# Samples'' is the number of test examples we used for robustness evaluations}
\label{tb:datsets}
\begin{center}
\begin{sc}
\begin{small}
\begin{tabular}{lccr}
\toprule
Datasets & \# samples & Average input lengths \\
\midrule
Gigaword    & 1,000& 30.1 words \\
DUC2003 & 624 & 35.5 words\\
DUC2004    & 500& 35.6 words \\
Multi30k & 500 & 11.5 words \\
\bottomrule
\end{tabular}
\end{small}
\end{sc}
\end{center}
\end{table}

\subsection{Seq2seq models}
We implement both text summarization and machine translation models on OpenNMT-py. Specifically, we use a word-level LSTM encoder and a word-based attention decoder for both applications~\citep{bahdanau2014neural}. For the text summarization task, we use 380k training pairs from Gigaword dataset to train a seq2seq model. 
The architecture consists of a 2-layer stacked LSTM with 500 hidden units. We conduct experiments on two types of models, one uses the pre-trained 300-dimensional GloVe word embeddings and the other one is trained from scratch. We set the beam search size to be 5 as suggested. For the machine translation task, we train our model using 453k pairs from the Europal corpus of German-English WMT 15
, common crawl and news-commentary. We use the hyper-parameters suggested by OpenNMT for both models, and have reproduced the performance reported in~\citet{rush2015neural} and \citet{ha2016toward}. 
\subsection{Empirical Results}
\subsubsection{Text Summarization}
For the non-overlapping attack, we use the proposed loss \eqref{eq:untargeted} in our objective function. 
A non-overlapping attack is treated as successful only if there is no common word at every position between output sequence and original sequence. We set $\lambda=1$ in all non-overlapping 
experiments. Table~\ref{sum:un} summarizes the experimental results. It shows that our algorithm only needs to change 2 or 3 words on average and can generate entirely different outputs for more than $80\%$ of sentences. 
We have also included some adversarial examples in Table ~\ref{nts:un}. 
From these examples, we can only change one word to let output sequence look completely different with the original one and change the sentence's meaning completely.
\begin{table}[htbp]
\caption{Results of non-overlapping attack in text summarization. {\bf \# changed} is how many words are changed in the input sentence. The high BLEU scores and low average number of changed words indicate that the crafted adversarial inputs are very similar to their originals, and we achieve high success rates to generate a summarization that differs with the original \textit{at every position} for all three datasets.}
\label{sum:un}
\begin{center}
\begin{small}
\begin{tabular}{lcccr}
\toprule
Dataset & Success\% & BLEU & \# changed \\
\midrule
Gigaword &86.0\% & 0.828 &2.17\\
DUC2003 &85.2\% & 0.774 &2.90 \\
DUC2004 &84.2\% & 0.816 &2.50 \\
\bottomrule
\end{tabular}
\end{small}
\end{center}
\end{table}

For the targeted keywords attack, we randomly choose some targeted keywords from the output vocabulary after removing the stop words like ``a'' and ``the''. 
A targeted keywords attack is treated as successful only if the output sequence contains all the targeted keywords. We set $\lambda_1=\lambda_2=1$ in our objective function~\eqref{eq:loss} in all our experiments. Table ~\ref{sum:num} summarizes the performance, including the overall success rate, average BLEU score ~\cite{papineni2002bleu}, and the average number of changed words in input sentences. Average BLEU score is defined by exponential average over BLEU 1,2,3,4, which is commonly used in evaluating the quality of text which has been machine-translated from one natural language to another. Also, we have included some adversarial examples crafted by our method in Table~\ref{sum:examples}. In Table~\ref{sum:examples}, some adversarial examples with 3 sets of keywords, where ``\#\#'' stands for a two-digit number after standard preprocessing in text summarization. Through these examples, our method could generate totally irrelevant subjects, verbs, numerals and objects which could easily be formed as a complete sentence with only several word changes. 
Note that there are three important techniques used in our algorithm: projected gradient method, group lasso, and gradient regularization. Therefore, we conduct experiments to verify the importance of each of these techniques. 

\begin{table}[h]
\caption{Results of targeted keywords attack in text summarization. $|K|$ is the number of keywords. We found that our method can make the summarization include 1 or 2 target keywords with a high success rate, while the changes made to the input sentences are relatively small, as indicated by the high BLEU scores and low average number of changed words. When $|K|=3$, this task becomes more challenging, but our algorithm can still find many adversarial examples.}
\label{sum:num}
\begin{center}
\begin{small}
\begin{tabular}{lccccr}
\toprule
Datasest & $|K|$ & Success\% & BLEU & \# changed\\
\midrule
\multirow{3}{*}{Gigaword} &1   &99.8\% & 0.801 &  2.04 &  \\ 
&2   & 96.5\% & 0.523 & 4.96 &\\
&3   &43.0\% & 0.413  & 8.86 &\\
\bottomrule
\toprule
\multirow{3}{*}{DUC2003} &1   &99.6\%  & 0.782 &  2.25\\ 
&2   &87.6\% & 0.457 & 5.57  &\\
&3   &38.3\% & 0.376  & 9.35 &\\
\bottomrule
\toprule
\multirow{3}{*}{DUC2004} &1    &99.6\% & 0.773 &  2.21 &\\ 
&2   &87.8\% & 0.421  & 5.1 & \\
&3   &37.4\% & 0.340  & 9.3 & \\
\bottomrule
\end{tabular}
\end{small}
\end{center}
\end{table}

\subsubsection{Machine Translation}
We then conduct both non-overlapping and targeted keywords attacks to the English-German machine translation model. We first filter out stop words like ``Ein''(a), ``und''(and) in German vocabulary and randomly choose several nouns, verbs, adjectives or adverbs in German as targeted keywords. Similar to the text summarization experiments, we set $\lambda_1=\lambda_2=1$ in our objective function. The success rates, BLEU scores, and the average number of words changed are reported in Table~\ref{nmt:num}, with some adversarial examples shown in Table~\ref{nmt:example}. 
\begin{table}[htbp]
\caption{Results of non-overlapping method and targeted keywords method in machine translation.} 
\label{nmt:num}
\begin{center}
\begin{small}
\begin{tabular}{lccccr}
\toprule
Method & Success\% & BLEU & \# changed \\
\midrule
Non-overlap  & 89.4\%& 0.349 & 3.5 \\
1-keyword    &100.0\% & 0.705 & 1.8 \\ 
2-keyword    & 91.0 \% & 0.303 & 4.0 \\
3-keyword    &69.6\% & 0.205  & 5.3 \\
\bottomrule
\end{tabular}
\end{small}
\end{center}
\end{table}

\begin{table}[htbp]
    \centering 
    \caption{Perplexity score for adversarial example}
    \begin{small}
    \begin{tabular}{ccc}
    \toprule
         &  DUC2003 & DUC2004\\
         \hline
Original &  102.02 & 121.09 \\

Non-overlap & 114.02 & 149.15 \\

1-keyword & 159.54 & 199.01\\

2-keyword & 352.12 & 384.80\\
\bottomrule
 \end{tabular}
       \label{tab:perplexity}
    \end{small}
\end{table}

\subsection{Analysis of Syntactic structure and Semantic Meaning Preservation}
In our algorithm we aim to make adversarial examples having similar meaning to original examples by constraining the number of changed words and enforcing the changed words are close to the original words in the embedding space. {However, depending on the implemented word embedding techniques, in general there is no guarantee that every word pair close in the embedding space have similar meanings.} Therefore, we have conducted additional experiments to verify the syntactic and semantic quality of our generated adversarial examples. For syntactic structure part, as showed in Table~\ref{tab:perplexity}, we measure the perplexity of generated adversarial sentences in DUC2003 and DUC2004 dataset. It shows that our examples keeps the original syntactic structure. For the semantic meaning part, We use DeepAI's online sentiment analysis {API} to test whether our attack changes the sentiment of 500 sentences from DUC2003 dataset in summarization task. The results show that 
{\bf only 2.2\% of adversarial examples have semantic meaning differ from the original sentences}. It proves that almost all adversarial examples keep the same semantic classification unchanged. 

\section{Analysis and Discussions}
\paragraph{Observation from adversarial example}
As shown in Table \ref{sum:examples}, our targeted keyword attack wouldn't just directly replace the keyword with some word in the source input. However, the word changed in the adversarial example and the target keyword are co-occurrent in the training dataset. It infers that seq2seq model learns the relationship between changed word and target keyword. However, the model fails to decide where it should focus on, which is strongly related with attention layer used in the model. It encourages us to use self-attention such as transformer~\cite{vaswani2017attention} instead to extract all the attentions between any two words.
When attacking subword transformer model, the target 1 keyword attack has 17\% lower success rate and 0.13 lower BLEU score. It shows transformer model has a greater adversarial robustness.
\paragraph{Robustness of Seq2Seq Model}
Although our algorithm can achieve very good success rates ($84\%-100\%$) 
in both non-overlapping and targeted keywords attacks with 1 or 2 keywords, we also recognize some strengths of the seq2seq model: (i) unlike CNN models where targeted attack can be conducted easily with almost 100\% success rate and very small distortion that cannot be perceived by human eyes~\cite{carlini2017towards}, it is harder to turn the entire seq2seq output into a particular sentence -- some sentences are even impossible to generate by seq2seq models; and (ii) since the input space of seq2seq is discrete, it is easier for human to detect the differences between the adversarial sequence and the original one, even if we only change one or few words. 
Therefore, we conclude that, compared with the DNN models designed for other tasks such as image classification, seq2seq models are more robust to adversarial attacks. The main reason, as pointed out in the introduction, is that the seq2seq model has a finite and discrete input space and 
almost infinite output space, so it is more robust than 
visual classification models that have an infinite and continuous input space and a very small output space (e.g., 10 categories in MNIST and 1,000 categories in ImageNet).

\begin{table*}[htbp]
\caption{Machine translation adversarial examples. Upper 4 lines: non-overlap; Bottom 4 lines: targeted keyword "Hund sitzt"}
\label{nmt:example}
\begin{center}
\begin{small}
\begin{tabular}{l|p{1.5\columnwidth}}
\toprule
Source input seq   & A child is splashing in the water.\\
\hline
Adv input seq & A children is {\bf unionists} in the water.\\
\hline
Source output seq   & Ein Kind im Wasser. \\
\hline
Adv output seq & {\bf Kinder sind in der Wasser @-@ $<$unk$>$}.
\\
\bottomrule
\toprule
Source input seq   & Two men wearing swim trunks jump in the air at a moderately populated beach.\\
\hline
Adv input seq & Two men wearing {\bf dog Leon comes} in the air at a moderately populated beach.\\
\hline
Source output seq   & Zwei Männer in Badehosen springen auf einem mäßig belebten Strand in die Luft.\\
\hline
Adv output seq & Zwei Männer tragen {\bf Hund} , der in der Luft {\bf sitzt} , hat $<$unk$>$ $<$unk$>$ .\\
\bottomrule
\end{tabular}
\end{small}
\end{center}
\end{table*}

\begin{table*}[htbp]
\caption{Text summarization adversarial examples using non-overlapping method. Surprisingly, it is possible to make the output sequence completely different by changing only one word in the input sequence.}
\label{nts:un}
\begin{center}
\begin{small}
\begin{tabular}{l|p{1.5\columnwidth}}
\toprule
Source input seq  & among asia 's leaders , prime minister mahathir mohamad was notable as a man with a bold vision : a physical and social transformation that would push this nation into the forefront of world affairs . \\
\hline
Adv input seq & among {\bf lynn} 's leaders , prime minister mahathir mohamad was notable as a man with a bold vision : a physical and social transformation that would push this nation into the forefront of world affairs.\\
\hline
Source output seq   & asia 's leaders are a man of the world\\
\hline
Adv output seq & {\bf a vision for the world}\\
\bottomrule
\toprule
Source input seq   & under nato threat to end his punishing offensive against ethnic albanian separatists in kosovo , president slobodan milosevic of yugoslavia has ordered most units of his army back to their barracks and may well avoid an attack by the alliance , military observers and diplomats say \\
\hline
Adv input seq & under nato threat to end his punishing offensive against ethnic albanian separatists in kosovo , president slobodan milosevic of yugoslavia has {\bf jean-sebastien} most units of his army back to their barracks and may well avoid an attack by the alliance , military observers and diplomats say.\\
\hline
Source output seq   & milosevic orders army back to barracks\\
\hline
Adv output seq & {\bf nato may not attack kosovo}\\
\bottomrule
\toprule
Source input seq &  flooding on the yangtze river remains serious although water levels on parts of the river decreased today , according to the state headquarters of flood control and drought relief .\\
\hline
Adv input seq & flooding {\bf that} the yangtze river {\bf becomes} serious although water levels on parts of the river decreased today , according to the state headquarters of flood control and drought relief . \\
\hline
Source output seq & floods on yangtze river continue \\
\hline
Adv output seq &  {\bf flooding in water recedes in river}  \\
\bottomrule
\end{tabular}
\end{small}
\end{center}
\end{table*}

\begin{table*}[htbp]
\caption{Text summarization examples with targeted keywords ``police arrest ''}
\label{sum:examples}
\begin{center}
\begin{small}
\begin{tabular}{l|p{1.5\columnwidth}}
\toprule
Source input seq   & north korea is entering its fourth winter of chronic food shortages with its people malnourished and at risk of dying from normally curable illnesses , senior red cross officials said tuesday. \\
\hline
Adv input seq & north {\bf detectives} is {\bf apprehended} its fourth winter of chronic food shortages with its people malnourished and at risk of dying from normally curable illnesses , senior red cross officials said tuesday.\\
\hline
Source output seq   &north korea enters fourth winter of food shortages\\
\hline
Adv output seq &  north {\bf police arrest} fourth winter of food shortages.\\
\bottomrule
\toprule
Source input seq   & after a day of fighting , congolese rebels said sunday they had entered kindu , the strategic town and airbase in eastern congo used by the government to halt their advances. \\
\hline
Adv input seq & after a day of fighting , {\bf nordic detectives} said sunday they had entered {\bf UNK} , the strategic town and airbase in eastern congo used by the government to halt their advances.\\
\hline
Source output seq   &congolese rebels say they have entered UNK.\\
\hline
Adv output seq &  nordic {\bf police arrest} \#\# in congo.\\
\bottomrule
\toprule
Source input seq   & president boris yeltsin stayed home tuesday , nursing a respiratory infection that forced him to cut short a foreign trip and revived concerns about his ability to govern. \\
\hline
Adv input seq & president boris yeltsin stayed home tuesday , {\bf cops cops} respiratory infection that forced him to cut short a foreign trip and revived concerns about his ability to govern.\\
\hline
Source output seq   & yeltsin stays home after illness\\
\hline
Adv output seq & yeltsin stays home after {\bf police arrest}\\
\bottomrule
\end{tabular}
\end{small}
\end{center}
\end{table*}

\section{Conclusion}

In this paper, we propose a novel framework, i.e., Seq2Sick, to generate adversarial examples for sequence-to-sequence neural network models. We propose a projected gradient method to address the issue of discrete input space, adopt group lasso to enforce the sparsity of the distortion, and develop a regularization technique to further improve the success rate. Besides, different from most existing algorithms that are designed for untargeted attack and classification tasks, our algorithm can perform the more challenging targeted keywords attack. Our experimental results show that the proposed framework is powerful and effective: it can achieve high success rates in both non-overlapping and targeted keywords attacks with relatively small distortions {and preserve similar sentiment classification results for the most of the generated adversarial examples}. 

\bibliographystyle{aaai}
\bibliography{aaai2020.bib}

\begin{thebibliography}{}

\bibitem[\protect\citeauthoryear{Alzantot \bgroup et al\mbox.\egroup
  }{2018}]{alzantot2018generating}
Alzantot, M.; Sharma, Y.; Elgohary, A.; Ho, B.-J.; Srivastava, M.; and Chang,
  K.-W.
\newblock 2018.
\newblock Generating natural language adversarial examples.
\newblock In {\em Proceedings of the 2018 Conference on Empirical Methods in
  Natural Language Processing},  2890--2896.

\bibitem[\protect\citeauthoryear{Bahdanau, Cho, and
  Bengio}{2014}]{bahdanau2014neural}
Bahdanau, D.; Cho, K.; and Bengio, Y.
\newblock 2014.
\newblock Neural machine translation by jointly learning to align and
  translate.
\newblock {\em arXiv preprint arXiv:1409.0473}.

\bibitem[\protect\citeauthoryear{Carlini and Wagner}{2017}]{carlini2017towards}
Carlini, N., and Wagner, D.
\newblock 2017.
\newblock Towards evaluating the robustness of neural networks.
\newblock In {\em Security and Privacy (SP), 2017 IEEE Symposium on},  39--57.
\newblock IEEE.

\bibitem[\protect\citeauthoryear{Chan \bgroup et al\mbox.\egroup
  }{2016}]{chan2016listen}
Chan, W.; Jaitly, N.; Le, Q.; and Vinyals, O.
\newblock 2016.
\newblock Listen, attend and spell: A neural network for large vocabulary
  conversational speech recognition.
\newblock In {\em 2016 IEEE International Conference on Acoustics, Speech and
  Signal Processing (ICASSP)},  4960--4964.
\newblock IEEE.

\bibitem[\protect\citeauthoryear{Ebrahimi \bgroup et al\mbox.\egroup
  }{2017}]{ebrahimi2017hotflip}
Ebrahimi, J.; Rao, A.; Lowd, D.; and Dou, D.
\newblock 2017.
\newblock Hotflip: White-box adversarial examples for nlp.
\newblock {\em arXiv preprint arXiv:1712.06751}.

\bibitem[\protect\citeauthoryear{Friedman}{1997}]{friedman1997bias}
Friedman, J.~H.
\newblock 1997.
\newblock On bias, variance, 0/1—loss, and the curse-of-dimensionality.
\newblock {\em Data mining and knowledge discovery} 1(1):55--77.

\bibitem[\protect\citeauthoryear{Gao \bgroup et al\mbox.\egroup
  }{2018}]{gao2018black}
Gao, J.; Lanchantin, J.; Soffa, M.~L.; and Qi, Y.
\newblock 2018.
\newblock Black-box generation of adversarial text sequences to evade deep
  learning classifiers.
\newblock {\em arXiv preprint arXiv:1801.04354}.

\bibitem[\protect\citeauthoryear{Gong \bgroup et al\mbox.\egroup
  }{2018}]{gong2018adversarial}
Gong, Z.; Wang, W.; Li, B.; Song, D.; and Ku, W.-S.
\newblock 2018.
\newblock Adversarial texts with gradient methods.
\newblock {\em arXiv preprint arXiv:1801.07175}.

\bibitem[\protect\citeauthoryear{Goodfellow, Shlens, and
  Szegedy}{2014}]{goodfellow2014explaining}
Goodfellow, I.~J.; Shlens, J.; and Szegedy, C.
\newblock 2014.
\newblock Explaining and harnessing adversarial examples.
\newblock {\em arXiv preprint arXiv:1412.6572}.

\bibitem[\protect\citeauthoryear{Ha, Niehues, and Waibel}{2016}]{ha2016toward}
Ha, T.-L.; Niehues, J.; and Waibel, A.
\newblock 2016.
\newblock Toward multilingual neural machine translation with universal encoder
  and decoder.
\newblock {\em arXiv preprint arXiv:1611.04798}.

\bibitem[\protect\citeauthoryear{Jia and Liang}{2017}]{jia2017adversarial}
Jia, R., and Liang, P.
\newblock 2017.
\newblock Adversarial examples for evaluating reading comprehension systems.
\newblock {\em arXiv preprint arXiv:1707.07328}.

\bibitem[\protect\citeauthoryear{Li, Monroe, and
  Jurafsky}{2016}]{li2016understanding}
Li, J.; Monroe, W.; and Jurafsky, D.
\newblock 2016.
\newblock Understanding neural networks through representation erasure.
\newblock {\em arXiv preprint arXiv:1612.08220}.

\bibitem[\protect\citeauthoryear{Liang \bgroup et al\mbox.\egroup
  }{2017}]{liang2017deep}
Liang, B.; Li, H.; Su, M.; Bian, P.; Li, X.; and Shi, W.
\newblock 2017.
\newblock Deep text classification can be fooled.
\newblock {\em arXiv preprint arXiv:1704.08006}.

\bibitem[\protect\citeauthoryear{Michel \bgroup et al\mbox.\egroup
  }{2019}]{michel2019evaluation}
Michel, P.; Li, X.; Neubig, G.; and Pino, J.~M.
\newblock 2019.
\newblock On evaluation of adversarial perturbations for sequence-to-sequence
  models.
\newblock {\em arXiv preprint arXiv:1903.06620}.

\bibitem[\protect\citeauthoryear{Papernot \bgroup et al\mbox.\egroup
  }{2016}]{papernot2016crafting}
Papernot, N.; McDaniel, P.; Swami, A.; and Harang, R.
\newblock 2016.
\newblock Crafting adversarial input sequences for recurrent neural networks.
\newblock In {\em Military Communications Conference, MILCOM 2016-2016 IEEE},
  49--54.
\newblock IEEE.

\bibitem[\protect\citeauthoryear{Papineni \bgroup et al\mbox.\egroup
  }{2002}]{papineni2002bleu}
Papineni, K.; Roukos, S.; Ward, T.; and Zhu, W.-J.
\newblock 2002.
\newblock Bleu: a method for automatic evaluation of machine translation.
\newblock In {\em Proceedings of the 40th annual meeting on association for
  computational linguistics},  311--318.
\newblock Association for Computational Linguistics.

\bibitem[\protect\citeauthoryear{Pennington, Socher, and
  Manning}{2014}]{pennington2014glove}
Pennington, J.; Socher, R.; and Manning, C.
\newblock 2014.
\newblock Glove: Global vectors for word representation.
\newblock In {\em Proceedings of the 2014 conference on empirical methods in
  natural language processing (EMNLP)},  1532--1543.

\bibitem[\protect\citeauthoryear{Rush, Chopra, and
  Weston}{2015}]{rush2015neural}
Rush, A.~M.; Chopra, S.; and Weston, J.
\newblock 2015.
\newblock A neural attention model for abstractive sentence summarization.
\newblock {\em arXiv preprint arXiv:1509.00685}.

\bibitem[\protect\citeauthoryear{Samanta and Mehta}{2017}]{samanta2017towards}
Samanta, S., and Mehta, S.
\newblock 2017.
\newblock Towards crafting text adversarial samples.
\newblock {\em arXiv preprint arXiv:1707.02812}.

\bibitem[\protect\citeauthoryear{Sutskever, Vinyals, and
  Le}{2014}]{DBLP:conf/nips/SutskeverVL14}
Sutskever, I.; Vinyals, O.; and Le, Q.~V.
\newblock 2014.
\newblock Sequence to sequence learning with neural networks.
\newblock In {\em Advances in Neural Information Processing Systems 27: Annual
  Conference on Neural Information Processing Systems 2014, December 8-13 2014,
  Montreal, Quebec, Canada},  3104--3112.

\bibitem[\protect\citeauthoryear{Szegedy \bgroup et al\mbox.\egroup
  }{2013}]{DBLP:journals/corr/SzegedyZSBEGF13}
Szegedy, C.; Zaremba, W.; Sutskever, I.; Bruna, J.; Erhan, D.; Goodfellow,
  I.~J.; and Fergus, R.
\newblock 2013.
\newblock Intriguing properties of neural networks.
\newblock {\em CoRR} abs/1312.6199.

\bibitem[\protect\citeauthoryear{Vaswani \bgroup et al\mbox.\egroup
  }{2017}]{vaswani2017attention}
Vaswani, A.; Shazeer, N.; Parmar, N.; Uszkoreit, J.; Jones, L.; Gomez, A.~N.;
  Kaiser, {\L}.; and Polosukhin, I.
\newblock 2017.
\newblock Attention is all you need.
\newblock In {\em Advances in neural information processing systems},
  5998--6008.

\bibitem[\protect\citeauthoryear{Yang \bgroup et al\mbox.\egroup
  }{2018}]{yang2018greedy}
Yang, P.; Chen, J.; Hsieh, C.-J.; Wang, J.-L.; and Jordan, M.~I.
\newblock 2018.
\newblock Greedy attack and gumbel attack: Generating adversarial examples for
  discrete data.
\newblock {\em arXiv preprint arXiv:1805.12316}.

\bibitem[\protect\citeauthoryear{Zhao, Dua, and
  Singh}{2017}]{zhao2017generating}
Zhao, Z.; Dua, D.; and Singh, S.
\newblock 2017.
\newblock Generating natural adversarial examples.
\newblock {\em arXiv preprint arXiv:1710.11342}.

\end{thebibliography}




\end{document}